\documentclass{sig-alternate-05-2015}
\usepackage{times}
\usepackage{latexsym}
\usepackage{url}
\usepackage{hyperref}
\usepackage{breakurl}
\usepackage{graphicx}
\usepackage[table]{xcolor}
\usepackage{stfloats}
\usepackage{pgfplots}
\usepackage{multicol}
\usepackage{balance}

\begin{document}






%

\title{Predicting the Industry of Users on Social Media}
%
%
%
%
%

\numberofauthors{2} 
%
\author{ 
%
%
\alignauthor 
Konstantinos Pappas \\ 
       \affaddr{University of Michigan}\\ 
       \affaddr{Computer Science and Engineering}\\ 
       \affaddr{Ann Arbor, MI 48109, USA}\\ 
       \email{pappus@umich.edu} 
\alignauthor 
Rada Mihalcea \\ 
       \affaddr{University of Michigan}\\ 
       \affaddr{Computer Science and Engineering}\\ 
       \affaddr{Ann Arbor, MI 48109, USA}\\ 
       \email{mihalcea@umich.edu} 
}

\maketitle
\begin{abstract}
Automatic profiling of social media users is an important task for supporting a multitude of downstream applications. While a number of studies have used social media content to extract and study collective social attributes, there is a lack of substantial research that addresses the detection of a user's industry. We frame this task as classification using both feature engineering and ensemble learning. Our industry-detection system uses both posted content and profile information to detect a user's industry with 64.3\% accuracy, significantly outperforming the majority baseline in a taxonomy of fourteen industry classes. Our qualitative analysis suggests that a person's industry not only affects the words used and their perceived meanings, but also the number and type of emotions being expressed.
\end{abstract}

%
%


%
%

%
%



\section{Introduction}
Over the past two decades, the emergence of social media has enabled the proliferation of traceable human behavior. The content posted by users can reflect who their friends are, what topics they are interested in, or which company they are working for. At the same time, users are listing a number of profile fields to define themselves to others. The utilization of such metadata has proven important in facilitating further developments of applications in advertising \cite{Bharat16}, personalization \cite{Fink00}, and recommender systems \cite{Adomavicius11}. However, profile information can be limited, depending on the platform, or it is often deliberately omitted \cite{Hernandez13}. To uncloak this information, a number of studies have utilized social media users' footprints to approximate their profiles.

This paper explores the potential of predicting a user's industry --the aggregate of enterprises in a particular field-- by identifying industry indicative text in social media. The accurate prediction of users' industry can have a big impact on targeted advertising by minimizing wasted advertising \cite{Johnson13} and improved personalized user experience. A number of studies in the social sciences have associated language use with social factors such as occupation, social class, education, and income \cite{Bernstein60,Labov72,Bernstein03,Labov06}. An additional goal of this paper is to examine such findings, and in particular the link between language and occupational class, through a data-driven approach.

In addition, we explore how meaning changes depending on the occupational context. By leveraging word embeddings, we seek to quantify how, for example, {\it cloud} might mean a separate concept (e.g., condensed water vapor) in the text written by users that work in environmental jobs while it might be used differently by users in technology occupations (e.g., Internet-based computing).

\begin{table*}[t]
\begin{center}
\scalebox{1}{
\begin{tabular}{ l r | l r }
\hline
\hline
Technology              & 4,175 & Law                                  & 1,520 \\
Religion                & 3,165 & Security/Military & 933 \\
Fashion                 & 2,119 & Tourism                              & 840 \\
Publishing              & 2,102 & Construction                         & 837 \\
Sports or Recreation    & 1,779 & Museums or Libraries                 & 823 \\
Real Estate             & 1,726 & Banking/Investment Banking           & 735 \\
Agriculture/Environment & 1,620 & Automotive                           & 506 \\
\hline
\hline
\end{tabular}
}
\caption{Industry categories and number of users per category.\label{tab:ind}}
\end{center}
\end{table*}

\begin{table*}[t]
\centering
\scalebox{1}{
\begin{tabular}{ l c c c c }
\hline
Data per User & max & mean & $\sigma$ & median \\
\hline \hline 
Blogs & 97 & 1.8 & 2.9 & 1 \\
Blog Posts & 1356 & 24.5 & 30.4 & 21\\
Characters & 4,939,258 & 56,948 & 112,048.1 & 33,404 \\
\hline
\end{tabular}
}
\caption{Statistics on the Blogger dataset.\label{table:statisticsBlogger}}
\vspace{-0.1in}
\end{table*}

Specifically, this paper makes four main contributions. First, we build a large, industry-annotated dataset that contains over 20,000 blog users. In addition to their posted text, we also link a number of user metadata including their gender, location, occupation, introduction and interests.

Second, we build content-based classifiers for the industry prediction task and study the effect of incorporating textual features from the users' profile metadata using various meta-classification techniques, significantly improving both the overall accuracy and the average per industry accuracy.

Next, after examining which words are indicative for each industry, we build vector-space representations of word meanings and calculate one deviation for each industry, illustrating how meaning is differentiated based on the users' industries. We qualitatively examine the resulting industry-informed semantic representations of words by listing the words per industry that are most similar to job related and general interest terms.

Finally, we rank the different industries based on the normalized relative frequencies of emotionally charged words (positive and negative) and, in addition, discover that, for both genders, these frequencies do not statistically significantly correlate with an industry's gender dominance ratio.

After discussing related work in Section \ref{sec:related}, we present the dataset used in this study in Section \ref{sec:dataset}. In Section \ref{sec:text} we evaluate two feature selection methods and examine the industry inference problem using the text of the users' postings. We then augment our content-based classifier by building an ensemble that incorporates several metadata classifiers. We list the most industry indicative words and expose how each industrial semantic field varies with respect to a variety of terms in Section \ref{sec:qual}. We explore how the frequencies of emotionally charged words in each gender correlate with the industries and their respective gender dominance ratio and, finally, conclude in Section \ref{sec:con}.

\break

\section{Related Work}
\label{sec:related}

Alongside the wide adoption of social media by the public, researchers have been leveraging the newly available data to create and refine models of users' behavior and profiling. There exists a myriad research that analyzes language in order to profile social media users. Some studies sought to characterize users' personality \cite{Oberlander06,Celli12}, while others sequenced the expressed emotions \cite{Gil13}, studied mental disorders \cite{Harman14}, and the progression of health conditions \cite{Kashyap14}.
At the same time, a number of researchers sought to predict the social media users' age and/or gender \cite{Schler06,Rao10,Kokkos14}, while others targeted and analyzed the ethnicity, nationality, and race of the users \cite{Go09,Rao11,Mohammady14}. One of the profile fields that has drawn a great deal of attention is the location of a user. Among others, Hecht et al. \shortcite{Hecht11} predicted Twitter users' locations using machine learning on nationwide and state levels. Later, Han et al. \shortcite{Han14} identified location indicative words to predict the location of Twitter users down to the city level.

As a separate line of research, a number of studies have focused on discovering the political orientation of users \cite{Rao10,Lampos13,Volkova14}. Finally, Li et al. \shortcite{Li14a} proposed a way to model major life events such as getting married, moving to a new place, or graduating. In a subsequent study, \cite{Li14b} described a weakly supervised information extraction method that was used in conjunction with social network information to identify the name of a user's spouse, the college they attended, and the company where they are employed.

The line of work that is most closely related to our research is the one concerned with understanding the relation between people's language and their industry. Previous research from the fields of psychology and economics have explored the potential for predicting one's occupation from  their ability to use math and verbal symbols \cite{French59} and the relationship between job-types and demographics \cite{Schmidt75}. More recently, Huang et al. \shortcite{Huang15} used machine learning to classify Sina Weibo users to twelve different platform-defined occupational classes highlighting the effect of homophily in user interactions. This work examined only users that have been verified by the Sina Weibo platform, introducing a potential bias in the resulting dataset. Finally, Preotiuc-Pietro et al. \shortcite{Preoctiuc15} predicted the occupational class of Twitter users using the Standard Occupational Classification (SOC) system, which groups the different jobs based on skill requirements. In that work, the data collection process was limited to only users that specifically mentioned their occupation in their  self-description in a way that could be directly mapped to a SOC occupational  class. The mapping between a substring of their self-description and a SOC occupational class was done manually. Because of the manual annotation step, their method was not scalable; moreover, because they identified the occupation class inside a user self-description, only a very small fraction of the Twitter users could be included (in their case, 5,191 users).

Both of these recent studies are based on micro-blogging platforms, which inherently restrict the number of characters that a post can have, and consequently the way that users can express themselves. 

Moreover, both studies used off-the-shelf occupational taxonomies (rather than self-declared occupation categories), resulting in classes that are either too generic (e.g., media, welfare and electronic are three of the twelve Sina Weibo categories), or too intermixed (e.g., an assistant accountant is in a different class from an accountant in SOC). To address these limitations, we investigate the industry prediction task in a large blog corpus consisting of over 20K American users, 40K web-blogs, and 560K blog posts.

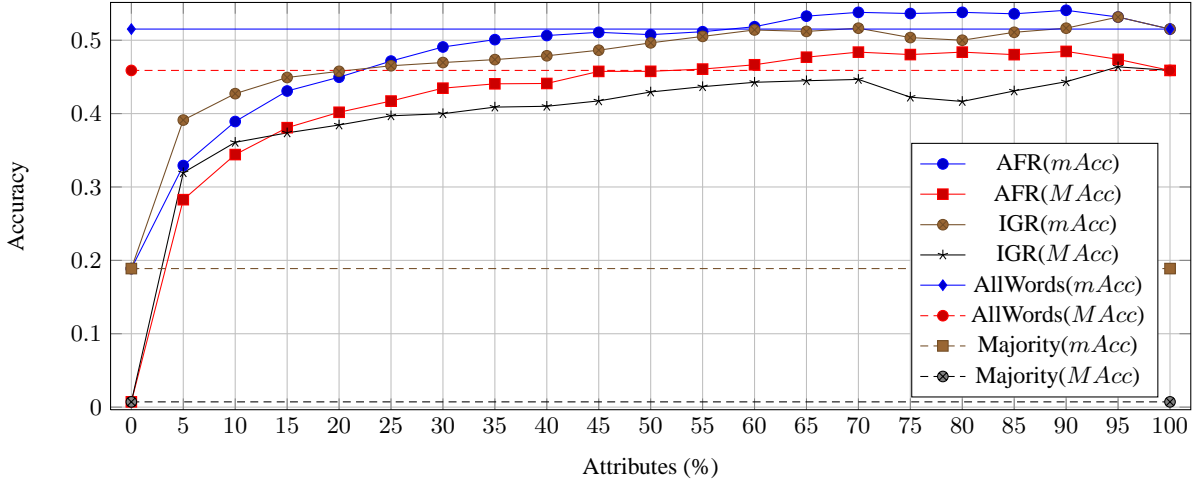
\begin{figure*}[t]
\begin{center}
\centering
\scalebox{1}{
\begin{tikzpicture}
\begin{axis}[
  legend pos=south east,
  height=7cm,
  width=0.9*\textwidth,
  enlargelimits=0.02,
  grid=major,
  xtick=data,
  ytick={0,0.1,0.2,0.3,0.4,0.5,0.6},
  ylabel={Accuracy},
  xlabel={Attributes (\%)}
  ]
  \addplot coordinates { (0,0.1888) (5,0.3292) (10,0.3892) (15,0.4308) (20,0.4496) (25,0.4716) (30,0.4908) (35,0.5008) (40,0.5064) (45,0.5108) (50,0.5076) (55,0.5116) (60,0.5184) (65,0.5328) (70,0.538) (75,0.5364) (80,0.538) (85,0.536) (90,0.5408) (95,0.5316) (100,0.5152) };
  \addlegendentry{AFR($mAcc$)}
  \addplot coordinates { (0,0.007142857142) (5,0.2827009515) (10,0.3441574326) (15,0.3808109847) (20,0.4018141885) (25,0.4171410154) (30,0.434662067) (35,0.4405955582) (40,0.4410775007) (45,0.4575156703) (50,0.4576266782) (55,0.4606991101) (60,0.4665416191) (65,0.4768797408) (70,0.483807679) (75,0.4804519227) (80,0.483807679) (85,0.480343261) (90,0.4847641869) (95,0.4739376886) (100,0.4588534261) };
  \addlegendentry{AFR($MAcc$)}
  \addplot coordinates { (0,0.1888) (5,0.3912) (10,0.4272) (15,0.4492) (20,0.4576) (25,0.4652) (30,0.4696) (35,0.4736) (40,0.4788) (45,0.4864) (50,0.4964) (55,0.5052) (60,0.514) (65,0.512) (70,0.5164) (75,0.5036) (80,0.5) (85,0.5108) (90,0.5164) (95,0.5316) (100,0.5152) };
  \addlegendentry{IGR($mAcc$)}
  \addplot coordinates { (0,0.007142857142) (5,0.3194606553) (10,0.3608456656) (15,0.3737478254) (20,0.3844726126) (25,0.3969896117) (30,0.3998707758) (35,0.4088637613) (40,0.409989415) (45,0.4173109352) (50,0.4294565372) (55,0.436701871) (60,0.442661769) (65,0.444803036) (70,0.4465807608) (75,0.4224322146) (80,0.4165546514) (85,0.4308769235) (90,0.4434525238) (95,0.4637295003) (100,0.4588534261) };
  \addlegendentry{IGR($MAcc$)}
  \addplot coordinates { (0,0.5152) (100,0.5152) };
  \addlegendentry{AllWords($mAcc$)}
  \addplot coordinates { (0,0.4588534261) (100,0.4588534261) };
  \addlegendentry{AllWords($MAcc$)}
  \addplot coordinates { (0,0.1888) (100,0.1888) };
  \addlegendentry{Majority($mAcc$)}
  \addplot coordinates { (0,0.007142857142) (100,0.007142857142) };
  \addlegendentry{Majority($MAcc$)}
\end{axis}
\end{tikzpicture}
}

\caption{Feature evaluation on the industry prediction task using Information Gain Ratio (IGR) and our Aggressive Feature Ranking (AFR). The performance is measured using both accuracy ($mAcc$) and average per-class accuracy ($MAcc$).\label{figure:featureSel}}
\end{center} 
\end{figure*}

\section{Dataset}
\label{sec:dataset}

We compile our industry-annotated dataset by identifying blogger profiles located in the U.S. on the profile finder on \url{http://www.blogger.com},  and scraping only those users that had the industry profile element completed.\footnote{This data collection was performed between May and July 2015.}  

For each of these bloggers, we retrieve all their blogs, and for each of these blogs we download  the  21  most  recent  blog  postings. We then clean these blog posts of HTML tags and tokenize them, and drop those  bloggers whose cumulative textual content in their posts is less than 600 characters. Following these guidelines, we identified all the U.S. bloggers with completed industry information. 

Traditionally, standardized industry taxonomies organize economic activities into groups based on similar production processes, products or services, delivery systems or behavior in financial markets. Following such assumptions and regardless of their many similarities, a tomato farmer would be categorized into a distinct industry from a tobacco farmer. As demonstrated in Preotiuc-Pietro et al. \shortcite{Preoctiuc15} such groupings can cause unwarranted misclassifications.

The Blogger platform provides a total of 39 different industry options. Even though a completed industry value is an implicit text annotation, we acknowledge the same problem noted in previous studies: some categories are too broad, while others are very similar. To remedy this and following Guibert et al. \shortcite{Guibert71}, who argued that the denominations used in a classification must reflect the purpose of the study, we group the different Blogger industries based on similar educational background and similar technical terminology. To do that, we exclude very general categories and merge conceptually similar ones\footnote{Merged categories are denoted with the '/' character in Table \ref{tab:ind}.}. Examples of broad categories are the {\it Education} and the {\it Student} options: a teacher could be teaching in any concentration, while a student could be enrolled in any discipline. Examples of conceptually similar categories are the {\it Investment Banking} and the {\it Banking} options.

The final set of categories is shown in Table \ref{tab:ind}, along with the number of users in each category. The resulting dataset consists of 22,880 users, 41,094 blogs, and 561,003 posts. Table \ref{table:statisticsBlogger} presents additional statistics of our dataset.

\section{Text-based Industry Modeling}
\label{sec:text}

After collecting our dataset, we split it into three sets: a train set, a development set, and a test set. The sizes of these sets are 17,880, 2,500, and 2,500 users, respectively, with users randomly assigned to these sets. In all the experiments that follow, we evaluate our classifiers by training them on the train set, configure the parameters and measure performance on the development set, and finally report the prediction accuracy and results on the test set. Note that all the experiments are performed at user level, i.e., all the data for one user is compiled into one instance in our data sets.

To measure the performance of our classifiers, we use the prediction accuracy. However, as shown in Table \ref{tab:ind}, the available data is skewed across categories, which could lead to somewhat distorted accuracy numbers depending on how well a model learns to predict the most populous classes. Moreover, accuracy alone does not provide a great deal of insight into the individual performance per industry, which is one of the main objectives in this study. Therefore, in our results below, we report: (1) micro-accuracy ($mAcc$), calculated as the percentage of correctly classified instances out of all the instances in the development (test) data; and (2) macro-accuracy ($MAcc$), calculated as the average of the per-category accuracies, where the per-category accuracy is the percentage of correctly classified instances out of the instances belonging to one category in the development (test) data.

\subsection{Leveraging Blog Content}

In this section, we seek the effectiveness of using solely textual features obtained from the users' postings to predict their industry. 

The industry prediction baseline {\it Majority} is set by discovering the most frequently featured class in our training set and picking that class in all predictions in the respective development or testing set. 

After excluding all the words that are not used by at least three separate users in our training set, we build our {\it AllWords} model by counting the frequencies of all the remaining words and training a multinomial Naive Bayes classifier. 
As seen in Figure \ref{figure:featureSel}, we can far exceed the {\it Majority} baseline performance by incorporating basic language signals into machine learning algorithms (173\% $mAcc$ improvement). 

We additionally explore the potential of improving our text classification task by applying a number of feature ranking methods and selecting varying proportions of top ranked features in an attempt to exclude noisy features. We start by ranking the different features, {\it w}, according to their Information Gain Ratio score (IGR) with respect to every industry, {\it i}, and training our classifier using different proportions of the top features.

\[\resizebox{0.48\textwidth}{!}{$IGR(w) = \frac{IG(w)}{IV(w)} \propto \frac{-H(i|w)}{-P(w)logP(w)-P(\overline{w})logP(\overline{w})} \propto $}\]

\[\resizebox{0.48\textwidth}{!}{$ \frac{P(w)\sum_{i \in I}P(i|w)logP(i|w)+P(\overline{w})\sum_{i \in I}P(i|\overline{w})logP(i|\overline{w})}{-P(w)logP(w)-P(\overline{w})logP(\overline{w})} $}\]

\begin{table*}[t]
\begin{center}
\scalebox{1}{
\begin{tabular}{ l c c c c c c}
\hline
\bf{Data} & Gender & Occupation & City & State & Introduction & Interests \\
\hline \hline
Train & 0.806 & 0.753 & 0.862 & 1.00 & 0.692 & 0.535 \\
Dev   & 0.814 & 0.712 & 0.788 & 1.00 & 0.671 & 0.549 \\
Test  & 0.812 & 0.709 & 0.768 & 1.00 & 0.686 & 0.533 \\
\hline
\end{tabular}
}
\caption{Proportion of users with non-empty metadata fields.\label{tab:metadata}}
\end{center}
\end{table*}

Even though we find that using the top 95\% of all the features already exceeds the performance of the {\it All Words} model on the development data, we further experiment with ranking our features with a more aggressive formula that heavily promotes the features that are tightly associated with any industry category. Therefore, for every word in our training set, we define our newly introduced ranking method, the Aggressive Feature Ranking (AFR), as:

\[AFR(w) = \max_{i \in I}{\frac{P(w|i)}{P(w)}}\]

In Figure \ref{figure:featureSel} we illustrate the performance of all four methods in our industry prediction task on the development data. Note that for each method, we provide both the accuracy ($mAcc$) and the average per-class accuracy ($MAcc$). The {\it Majority} and {\it All Words} methods apply to all the features; therefore, they are represented as a straight line in the figure. The {\it IGR} and {\it AFR} methods are applied to varying subsets of the features using a 5\% step.

Our experiments demonstrate that the word choice that the users make in their posts correlates with their industry. The first observation in Figure \ref{figure:featureSel} is that the $mAcc$ is proportional to $MAcc$; as $mAcc$ increases, so does $MAcc$. Secondly, the best result on the development set is achieved by using the top 90\%  of the features using the {\it AFR} method. Lastly, the improvements of the {\it IGR} and {\it AFR} feature selections are not substantially better in comparison to {\it All Words} (at most 5\% improvement between {\it All Words} and {\it AFR}), which suggest that only a few noisy features exist and most of the words play some role in shaping the ``language" of an industry.

As a final evaluation, we apply on the test data the classifier found to work best on the development data ({\it AFR} feature selection, top 90\% features), for an $mAcc$ of 0.534 and $MAcc$ of 0.477.

\subsection{Leveraging User Metadata}

Together with the industry information and the most recent postings of each blogger, we also download a number of accompanying profile elements. Using these additional elements, we explore the potential of incorporating users' metadata in our classifiers.

Table \ref{tab:metadata} shows the different user metadata we consider together with their coverage percentage (not all users provide a value for all of the profile elements). With the exception of the gender field, the remaining metadata elements shown in Table \ref{tab:metadata} are completed by the users as a freely editable text field. This introduces a considerable amount of noise in the set of possible metadata values. Examples of noise in the occupation field include values such as ``Retired'', ``I work.'', or ``momma'' which are not necessarily informative for our industry prediction task.

To examine whether the metadata fields can help in the prediction of a user's industry, 
we build classifiers using the different metadata elements. For each metadata element that has a textual value, we use all the words in the training set for that field as features. The only two exceptions are the {\it state} field, which is encoded as one feature that can take one out of 50 different values representing the 50 U.S. states; and the {\it gender} field, which is encoded as a feature with a distinct value for each user gender option: undefined, male, or female.

As shown in Table \ref{tab:individual}, we build four different classifiers using the multinomial NB algorithm: {\sc Occu} (which uses the words found in the {\it occupation} profile element), {\sc Intro} ({\it introduction}), {\sc Inter} ({\it interests}), and {\sc Gloc} (combined {\it gender}, {\it city}, {\it state}).

In general, all the metadata classifiers perform better than our majority baseline ($mAcc$ of 18.88\%). For the {\sc Gloc} classifier, this result is in alignment with previous studies \cite{Schmidt75}. However, the only metadata classifier that outperforms the content classifier is the {\sc Occu} classifier, which despite missing and noisy {\it occupation} values exceeds the content classifier's performance by an absolute 3.2\%. 

To investigate the promise of combining the five different classifiers we have built so far, we calculate their inter-prediction agreement using Fleiss's Kappa \cite{Fleiss71}, as well as the lower prediction bounds using the double fault measure \cite{Giacinto01}. The Kappa values, presented in the lower left side of Table \ref{tab:kappa}, express the classification agreement for categorical items, in this case the users' industry. Lower values, especially values below 30\%, mean smaller agreement. Since all five classifiers have better-than-baseline accuracy, this low agreement suggests that their predictions could potentially be combined to achieve a better accumulated result.

\begin{table}
\begin{center}
\scalebox{1}{
\begin{tabular}{ l c c }
\hline
Classifier & $mAcc$ & $MAcc$ \\
\hline \hline
{\sc Occu}  & {\bf 0.566} & {\bf 0.431} \\
{\sc Intro}  & 0.406 & 0.247 \\
{\sc Inter} & 0.287 & 0.157 \\
{\sc Gloc}  & 0.199 & 0.090 \\
\hline
\end{tabular}
}
\caption{Accuracy ($mAcc$) and average per-class accuracy ($MAcc$) of the base metadata classifiers on the development set.\label{tab:individual}}
\end{center}
\end{table}

\begin{table}
\begin{center}
\scalebox{1}{
\begin{tabular}{  c | c | c | c | c  }
\hline
\cellcolor[HTML]{AAAAAA}\color{black}{\sc Text} & 0.245 & 0.338 & 0.357 & 0.366 \\
\hline
\cellcolor[HTML]{DDDDDD}0.270 & \cellcolor[HTML]{AAAAAA}\color{black}{\sc Occu} & 0.348 & 0.386 & 0.409 \\
\hline
\cellcolor[HTML]{DDDDDD}0.186 & \cellcolor[HTML]{DDDDDD}0.303 & \cellcolor[HTML]{AAAAAA}\color{black}{\sc Text} & 0.535 & 0.554 \\
\hline
\cellcolor[HTML]{DDDDDD}0.019 & \cellcolor[HTML]{DDDDDD}0.153 & \cellcolor[HTML]{DDDDDD}0.216 & \cellcolor[HTML]{AAAAAA}\color{black}{\sc Inter} & 0.668 \\
\hline
\cellcolor[HTML]{DDDDDD}-0.129 & \cellcolor[HTML]{DDDDDD}-0.005 & \cellcolor[HTML]{DDDDDD}0.005 & \cellcolor[HTML]{DDDDDD}0.020 & \cellcolor[HTML]{AAAAAA}\color{black}{\sc Gloc} \\
\hline
\end{tabular}
}
\caption{Kappa scores and double fault results of the base classifiers on development data.\label{tab:kappa}}
\end{center}
\end{table}

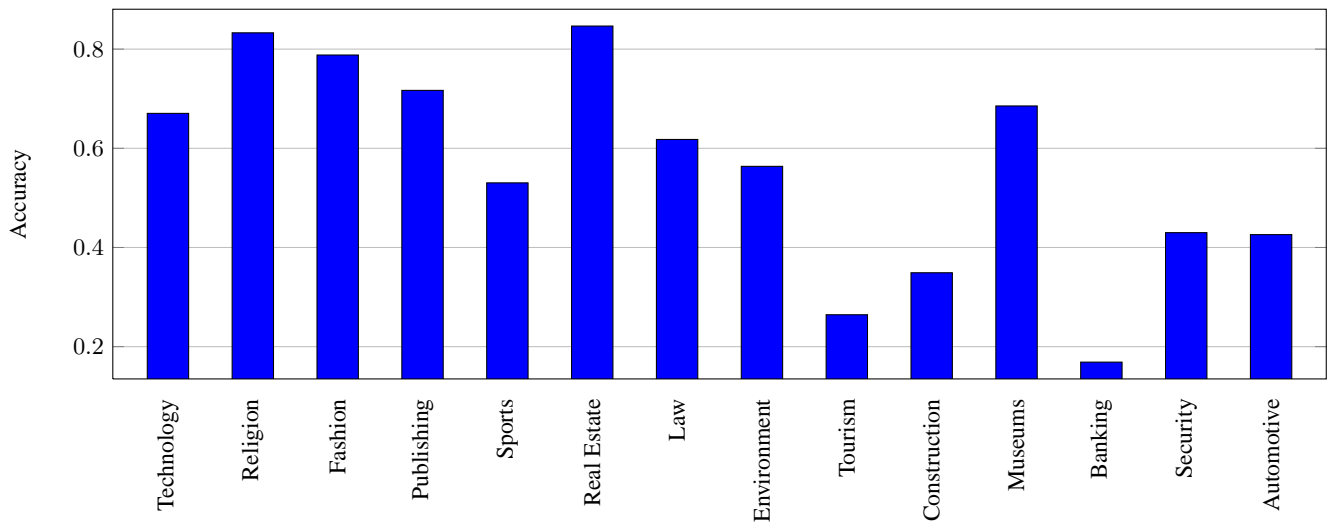
\begin{figure*}[b]
\begin{center}
\scalebox{1}{
\begin{tikzpicture}[y=5cm]
\begin{axis}[
  bar width = 0.55cm,
  height=6.5cm,
  width=1*\textwidth,
  major x tick style = transparent,
  ybar,
  enlargelimits=0.05,
  ylabel={Accuracy},
  symbolic x coords={Technology,Religion,Fashion,Publishing,Sports,Real Estate,Law,Environment,Tourism,Construction,Museums,Banking,Security,Automotive},
  xtick=data,
  x tick label style={rotate=90,anchor=east},
  ymajorgrids = true
  ]
\addplot[style={black,fill=blue,mark=none}] coordinates {(Technology,0.6704288939) (Religion,0.8328611898) (Fashion,0.7881355932) (Publishing,0.7168141593) (Sports,0.5303030303) (Real Estate,0.8465346535) (Law,0.6178343949) (Environment,0.5636363636) (Tourism,0.2643678161) (Construction,0.3490566038) (Museums,0.6853932584) (Banking,0.1688311688) (Security,0.4299065421) (Automotive,0.4259259259)};
\end{axis}
\end{tikzpicture}
}
\caption{Accuracy per-class using stacking meta-classification.\label{figure:ACA}}
\label{fig:ACA}
\end{center} 
\end{figure*}

\begin{table*}[t]
\begin{center}
\scalebox{1}{
\begin{tabular}{ l c c c c }
\hline
 & \multicolumn{2}{c}{Feature Concatenation} & \multicolumn{2}{c}{Stacking} \\
\cline{2-5}
Meta-classifiers & $mAcc$ & $MAcc$ & $mAcc$ & $MAcc$ \\
\hline \hline 
1. {\sc Text} + {\sc Occu} & 0.545 & 0.489 & 0.640 & 0.557 \\
2. \{1\} + {\sc Intro} & 0.546 & 0.487 & 0.648 & 0.560 \\
3. \{2\} + {\sc Inter} & 0.546 & 0.482 & {\bf 0.653} & {\bf 0.569} \\
4. \{3\} + {\sc Gloc} & 0.545 & 0.478 & 0.650 & 0.566 \\

\hline
\end{tabular}
}
\caption{Performance of feature concatenation (early fusion) and stacking (late fusion)  on the development set.\label{tab:concat}}
\end{center}
\end{table*}

Moreover, the double fault measure values, which are presented in the top-right hand side of Table \ref{tab:kappa}, express the proportion of test cases for which both of the two respective classifiers make false predictions, essentially providing the lowest error bound for the pairwise ensemble classifier performance. The lower those numbers are, the greater the accuracy potential of any meta-classification scheme that combines those classifiers. Once again, the low double fault measure values suggest potential gain from a combination of the base classifiers into an ensemble of models.

After establishing the promise of creating an ensemble of classifiers, we implement two meta-classification approaches. First, we combine our classifiers using features concatenation (or early fusion). Starting with our content-based classifier ({\sc Text}), we successively add the features derived from each metadata element. The results, both micro- and macro-accuracy, are presented in Table \ref{tab:concat}. Even though all these four feature concatenation ensembles outperform the content-based classifier in the development set, they fail to outperform the {\sc Occu} classifier.

Second, we explore the potential of using stacked generalization (or late fusion) \cite{wolpert92}. The base classifiers, referred to as L0 classifiers, are trained on different folds of the training set and used to predict the class of the remaining instances. Those predictions are then used together with the true label of the training instances to train a second classifier, referred to as the L1 classifier: this L1 is used to produce the final prediction on both the development data and the test data.  Traditionally, stacking uses different machine learning algorithms on the same training data. However in our case, we use the same algorithm (multinomial NB) on heterogeneous data (i.e., different types of data such as content, occupation, introduction, interests, gender, city and state) in order to exploit all available sources of information.

The ensemble learning results on the development set are shown in Table \ref{tab:concat}. We notice a constant improvement for both metrics when adding more classifiers to our ensemble except for the {\sc Gloc} classifier, which slightly reduces the performance. The best result is achieved using an ensemble of the {\sc Text}, {\sc Occu}, {\sc Intro}, and {\sc Inter} L0 classifiers; the respective performance on the test set is an $mAcc$ of 0.643 and an $MAcc$ of 0.564. 

Finally, we present in Figure \ref{figure:ACA} the prediction accuracy for the final classifier for each of the different industries in our test dataset. Evidently, some industries are easier to predict than others. For example, while the {\it Real Estate} and {\it Religion} industries achieve accuracy figures above 80\%, other industries, such as the {\it Banking} industry, are predicted correctly in less than 17\% of the time. Anecdotal evidence drawn from the examination of the confusion matrix does not encourage any strong association of the {\it Banking} class with any other. The misclassifications are roughly uniform across all other classes, suggesting that the users in the {\it Banking} industry use language in a non-distinguishing way.

\section{Qualitative Analysis}
\label{sec:qual}

In this section, we provide a qualitative analysis of the language of the different industries. 

\subsection{Top-Ranked Words}

To conduct a qualitative exploration of which words indicate the industry of a user, Table \ref{tab:premier} shows the three top-ranking content words for the different industries using the {\it AFR} method.

\begin{table}[h]
\begin{center}
\scalebox{1}{
\begin{tabular}{ l | c }
\hline
Industry & Top-Ranked Words \\
\hline
Technology & software, file, data \\ 
Religion & ministry, jesus, pastor \\ 
Fashion & fashion, dress, hair \\ 
Publishing  & writers, novel, writer \\ 
Sports & coach, weight, exercise \\ 
Real Estate & estate, details, homes \\ 
Law & court, trial, agreement \\ 
Environment & farm, plants, plant \\ 
Tourism & guests, travel, hotel \\ 
Construction & roof, construction, union \\
Museums & library, museum, novel \\ 
Banking & secret, agent, bank \\ 
Security & officer, army, military \\
Automotive & vehicle, cars, insurance \\
\hline
\end{tabular}
}
\caption{Three top-ranked words for each industry.\label{tab:premier}}
\end{center}
\end{table}

Not surprisingly, the top ranked words align well with what we would intuitively expect for each industry. Even though most of these words are potentially used by many users regardless of their industry in our dataset, they are still distinguished by the {\it AFR} method because of the different frequencies of these words in the text of each industry.

\vspace{-0.2cm}
\subsection{Industry-specific Word Similarities}
\vspace{-0.1cm}

Next, we examine how the meaning of a word is shaped by the context in which it is uttered. In particular, we qualitatively investigate how the speakers' industry affects meaning by learning vector-space representations of words that take into account such contextual information. To achieve this, we apply the contextualized word embeddings proposed by Bamman et al. \shortcite{Bamman14}, which are based on an extension of the ``skip-gram" language model \cite{Mikolov13}. 

\begin{table}[ht]
\begin{center}
\scalebox{1}{
 \begin{tabular}{  l | c || l | c } 
 \multicolumn{2}{c}{Technology} & \multicolumn{2}{c}{Tourism} \\
 \hline
 term & cosine & term & cosine \\ [0.5ex] 
 \hline\hline
 customers & 1.000 & customers & 1.000 \\ 
 clients & 0.870 & guests & 0.816 \\
 consumers & 0.858 & opportunities & 0.789 \\
 companies & 0.832 & clients & 0.783 \\
 employees & 0.822 & itineraries & 0.778 \\ 
 users & 0.820 & choices & 0.769 \\
 developers & 0.818 & patrons & 0.767 \\
 providers & 0.817 & employees & 0.760 \\
 businesses & 0.813 & projects & 0.757 \\ 
 customer & 0.811 & provide & 0.753 \\ 
 \hline
\end{tabular}
}
\caption{Terms with the highest cosine similarity to the term {\it customers}.\label{tab:customers}}
\end{center}
\end{table}

In addition to learning a global representation for each word, these contextualized embeddings compute one deviation from the common word embedding representation for each contextual variable, in this case, an industry option. These deviations capture the terms' meaning variations (shifts in the $k$-dimensional space of the representations, where $k=100$ in our experiments) in the text of the different industries, however all the embeddings are in the same vector space to allow for comparisons to one another.

\begin{table}[!]
\vspace{-0.5cm}
\begin{center}
\scalebox{1}{
 \begin{tabular}{ l | c || l | c } 
 \multicolumn{2}{c}{Environment} & \multicolumn{2}{c}{Tourism} \\
 \hline
 term & cosine & term & cosine \\ [0.5ex] 
 \hline\hline
 food & 1.000 & food & 1.000 \\ 
 local & 0.824 & delicious & 0.843 \\
 produce & 0.812 & treats & 0.822 \\
 meat & 0.807 & pastries & 0.814 \\
 wholesome & 0.805 & sandwiches & 0.808 \\ 
 processed & 0.785 & burgers & 0.806 \\
 consumers & 0.777 & dishes & 0.801 \\
 meals & 0.774 & selections & 0.796 \\
 nutritionally & 0.774 & eating & 0.792 \\ 
 locally & 0.765 & hamburgers & 0.791 \\ 
 \hline
\end{tabular}
}
\caption{Terms with the highest cosine similarity to the term {\it food}.\label{tab:food}}
\end{center}
\end{table}

Using the word representations learned for each industry, we present in Table \ref{tab:customers} the terms in the {\it Technology} and the {\it Tourism} industries that have the highest cosine similarity with a job-related word, {\it customers}. Similarly, Table \ref{tab:food} shows the words in the {\it Environment} and the {\it Tourism} industries that are closest in meaning to a general interest word, {\it food}. More examples are given in the  Appendix \ref{sec:condif}.

The terms that rank highest in each industry are noticeably different. For example, as seen in Table \ref{tab:food}, while {\it food} in the {\it Environment} industry is similar to {\it nutritionally} and {\it locally}, in the {\it Tourism} industry the same word relates more to terms such as {\it delicious} and {\it pastries}. These results not only emphasize the existing differences in how people in different industries perceive certain terms, but they also demonstrate that those differences can effectively be captured in the resulting word embeddings. 

\vspace{-0.2cm}
\subsection{Emotional Orientation per Industry and Gender}

\vspace{-0.1cm}
As a final analysis, we explore how words that are emotionally charged relate to different industries. 
To quantify the emotional orientation of a text, we use the {\it Positive Emotion} and {\it Negative Emotion} categories in the Linguistic Inquiry and Word Count (LIWC) dictionary \cite{Tausczik10}. The LIWC dictionary contains lists of words that have been shown to correlate with the psychological states of people that use them; for example, the {\it Positive Emotion} category contains words such as ``happy,'' ``pretty,'' and ``good.''

For the text of all the users in each industry we measure the frequencies of {\it Positive Emotion} and {\it Negative Emotion} words  normalized by the text's length. Table \ref{tab:indRank} presents the industries' ranking for both categories of words based on their relative frequencies in the text of each industry.

We further perform a breakdown per-gender, where we  once again calculate the proportion of emotionally charged words in each industry, but separately for each gender. We find that the industry rankings of the relative frequencies $f_i$ of emotionally charged words for the two genders are statistically significantly correlated,\footnote{$\rho>$0.81 and p$<$0.001 for both categories of words.} which suggests that regardless of their gender, users use positive (or negative) words with a relative frequency that correlates with their industry. (In other words, even if e.g., {\it Fashion} has a larger number of women users, both men and women working in {\it Fashion} will tend to use more positive words than the corresponding gender in another industry with a larger number of men users such as {\it Automotive}.)

Finally, motivated by previous findings of correlations between job satisfaction and gender dominance in the workplace \cite{Bender05}, we explore the relationship between the usage of {\it Positive Emotion} and {\it Negative Emotion} words and the gender dominance in an industry. Although we find that there are substantial gender imbalances in each industry (Appendix \ref{sec:gendom}), we did not find any statistically significant correlation between the gender dominance ratio in the different industries and the usage of positive (or negative) emotional words in either gender in our dataset.

\begin{table}
\begin{center}
\scalebox{1}{
\begin{tabular}{ l | c || l| c  }
\hline
Positive & $f_i\times10^3$ & Negative & $f_i\times10^3$ \\
\hline
\hline
Fashion & 35.93 & Security & 13.80 \\
Religion & 32.10 & Religion & 13.68 \\
Tourism & 30.61 & Law & 12.97 \\
Banking & 30.44 & Publishing & 12.66 \\
Sports & 30.05 & Construction & 11.77 \\
Real Estate & 29.25 & Banking & 11.74 \\
Publishing & 29.12 & Sports & 10.68 \\
Security & 28.92 & Technology & 10.65 \\
Construction & 28.84 & Museums & 10.55 \\
Museums & 28.82 & Automotive & 10.53 \\
Environment & 28.31 & Environment & 10.17 \\
Law & 27.63 & Tourism & 9.53 \\
Automotive & 27.17 & Fashion & 8.50 \\
Technology & 26.42 & Real Estate & 8.25 \\
\hline
\end{tabular}
}
\caption{Ranking of industries based on the relative frequencies $f_i$ of {\it Positive Emotion} and {\it Negative Emotion} words.\label{tab:indRank}}
\end{center}
\end{table}

\vspace{-0.2cm}
\section{Conclusion}
\label{sec:con}

\vspace{-0.15cm}
In this paper, we examined the task of predicting a social media user's industry. We introduced an annotated dataset of over 20,000 blog users and applied a content-based classifier in conjunction with two feature selection methods for an overall accuracy of up to 0.534, which represents a large improvement over the majority class baseline of 0.188. 

We also demonstrated how the user metadata can be incorporated in our classifiers. Although concatenation of features drawn both from blog content and profile elements did not yield any clear improvements over the best individual classifiers, we found that stacking improves the prediction accuracy to an overall accuracy of 0.643, as measured on our test dataset. A more in-depth analysis showed that not all industries are equally easy to predict: while industries such as {\it Real Estate} and {\it Religion} are clearly distinguishable with accuracy figures over 0.80, others such as {\it Banking} are much harder to predict.

Finally, we presented a qualitative analysis to provide some insights into the language of different industries, which highlighted differences in the top-ranked words in each industry, word semantic similarities, and the relative frequency of emotionally charged words.

\section{Acknowledgments}
This material is based in part upon work supported by the National Science Foundation (\#1344257) and by the John Templeton Foundation (\#48503). Any opinions, findings, and conclusions or recommendations expressed in this material are those of the authors and do not necessarily reflect the views of the National Science Foundation or the John Templeton Foundation.

%
\bibliographystyle{abbrv}
\bibliography{sigproc}  
%
%
\appendix
\section{Additional Examples of Word Similarities}
\label{sec:condif}

\begin{table}[h!]
\begin{center}
\scalebox{1}{
 \begin{tabular}{ l | c || l | c } 
 \multicolumn{2}{c}{Religion} & \multicolumn{2}{c}{Sports} \\
 \hline
 term & cosine & term & cosine \\ [0.5ex] 
 \hline\hline
 professional & 1.000 & professional & 1.000 \\ 
 mentoring & 0.774 & sports & 0.833 \\
 education & 0.745 & coaching & 0.801 \\
 niche & 0.724 & active & 0.795 \\
 conversational & 0.722 & competitive & 0.793 \\ 
 vocational & 0.721 & becoming & 0.789 \\
 learner & 0.720 & major & 0.785 \\
 educational & 0.714 & fellow & 0.778 \\
 lock-ins & 0.714 & having & 0.775 \\ 
 thorough & 0.713 & coaches & 0.768 \\ 
 \hline
\end{tabular}
}
\caption{Terms with the highest cosine similarity to the term {\it professional}.\label{tab:professional}}
\end{center}
\end{table}


\vspace{-15pt}
\begin{table}[h!]
\begin{center}
\scalebox{1}{
 \begin{tabular}{ l | c || l | c } 
 \multicolumn{2}{c}{Technology} & \multicolumn{2}{c}{Fashion} \\
 \hline
 term & cosine & term & cosine \\ [0.5ex] 
 \hline\hline
 leisure & 1.000 & leisure & 1.000 \\ 
 playrooms & 0.651 & presale & 0.752 \\
 photo-editing & 0.650 & versona & 0.750 \\
 multi-media & 0.647 & jewerly & 0.748 \\
 match-making & 0.646 & high-end & 0.748 \\ 
 pre-ordered & 0.644 & sketchers & 0.747 \\
 tradeshows & 0.643 & craft & 0.743 \\
 tfp & 0.643 & vintage-inspired & 0.738 \\
 schmooze & 0.641 & spruill & 0.737 \\ 
 upload/download & 0.640 & baggu & 0.733 \\ 
 \hline
\end{tabular}
}
\caption{Terms with the highest cosine similarity to the term {\it leisure}.\label{tab:leisure}}
\end{center}
\end{table}

\vspace{-20pt}
\section{Gender Dominance in Industries}
\label{sec:gendom}

\begin{figure}[h!]
\begin{center}

\scalebox{1}{
\begin{tikzpicture}
\begin{axis}[
  symbolic y coords={Technology,Religion,Fashion,Publishing,Sports,Real Estate,Law,Environment,Tourism,Construction,Museums,Banking,Security,Automotive},
  xticklabels={1,0.8,0.6,0.4,0.2,0,0.2,0.4,0.6},
  ytick=data,
  grid=major
]
\addplot+[xbar] coordinates {(0.4832325454,Technology) (0.4497681607,Religion) (-0.807890223,Fashion) (-0.1546505228,Publishing) (0.1724659607,Sports) (-0.06311787072,Real Estate) (-0.09419152276,Law) (-0.2013769363,Environment) (-0.261589404,Tourism) (0.4028892456,Construction) (-0.5328719723,Museums) (-1.55E-03,Banking) (0.2716049383,Security) (0.568627451,Automotive)};

\end{axis}
\end{tikzpicture}
}

\hspace{2cm} Female \hspace{1cm} Male
\caption{Gender dominance for the different industries.\label{figure:genderImb}}

\end{center} 
\end{figure}
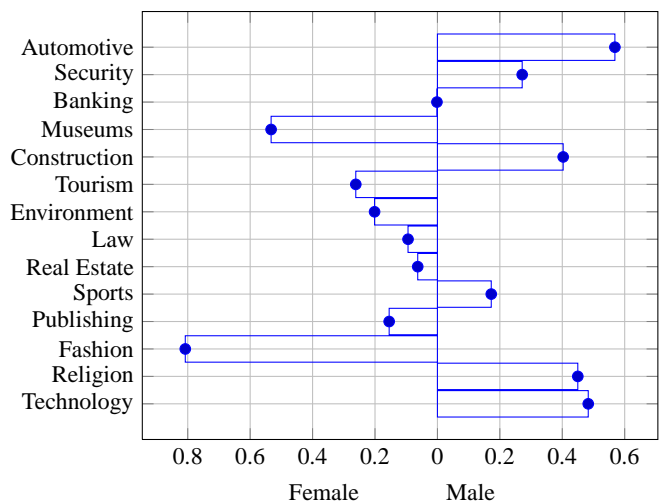

\end{document}